Paper ID 256

# 5G Features and Standards for Vehicle Data Exploitation

G. Velez[1*], E. Bonetto[2], D. Brevi[2], A. Martin[1], G. Rizzi[3], O. Castañeda[4],
A. Hamza-Cherif[5], M. Nieto[1], O. Otaegui[1]

1. {gvelez, amartin, mnieto, ootaegui}@vicomtech.org, Vicomtech Foundation, Spain

2. {edoardo.bonetto, daniele.brevi}@linksfoundation.com, Links Foundation, Italy

3. gianluca.rizzi@windtre.it, Wind Tre, Italy

4. oscaragustin.castaneda@dekra.com, Dekra, Spain

5. arslane.hamzacherif@{unimore, icoor}.it, UNIMORE & ICOOR, Italy

**Abstract**

Cars capture and generate huge volumes of data in real-time about the driving dynamics, the environment, and the driver and passengers' activities. Due to the proliferation of cooperative, connected and automated mobility (CCAM), the value of data from vehicles is getting strategic, not just for the automotive industry, but also for many diverse stakeholders including small and medium-sized enterprises (SMEs) and start-ups. 5G can enable car-captured data to feed innovative applications and services deployed in the cloud ensuring lower latency and higher throughput than previous cellular technologies. This paper identifies and discusses the relevance of the main 5G features that can contribute to a scalable, flexible, reliable and secure data pipeline, pointing to the standards and technical reports that specify their implementation.

## Introduction

The growing amount of available car data brings new opportunities around the automotive industry. For industry players, the exploitation of car data is strategic to achieve revenue generation, cost reduction and enhancement of safety and security. Tech Giants such as Google, Amazon or Microsoft have also targeted the global connected car data ecosystem [1]. OEMs and TIER1 suppliers are under pressure to develop their own data-driven value propositions by leveraging digital ecosystems and enforcing cooperation to fully influence data-driven business opportunities beyond autonomous driving and connected mobility domains, not only to improve existing processes and functions but also to enable entirely new business models. Extracting value from vehicle data will require new players in the automotive ecosystem.

High-tech SMEs and start-ups will become key players in the car data monetization landscape bringing free inspiration creativity, agile and quick development of digital innovations and realistic/pragmatic business scale. To this end, openness in terms of car data access is required allowing access to third parties.

5G will support the foreseen mobile data growth, and at the same time will allow for different new services classes. It is expected to bring network performance enhancements and agility in the network characteristics, by allowing the networks to be programmed and configured for specific use cases. Indeed, as introduced in 3GPP TS 28.530 [13], it is possible to create, over a common infrastructure, multiple logical network partitions (a.k.a. slices) that are properly isolated and that have allocated resources and topology optimized with respect to the use case of interest. By enabling and facilitating some of the major upcoming ICT trends, 5G will have an important role in supporting the exploitation of vehicle data by third parties.

This paper surveys the 5G features and standards that would be required by a data platform that aims to capture data from vehicles to make it available to third parties in the Cloud. The rest of the paper is organised as follows. Section 2 introduces the vehicle data exploitation platform. Section 3 describes the vehicular data space that needs to be considered by a vehicle data exploitation platform. Section 3 presents an overview of 5G features and standards. Section 5 describes the relevance of each of the identified 5G features for building a vehicle data platform. Section 6 presents the conclusions.

**Vehicle data exploitation platform**

This paper studies the 5G features and standards that would be necessary for building a platform that gives access to third parties to vehicle and Roadside Unit (RSU) data. Thus, this platform would connect vehicles and RSUs (data producers) with third-party applications and services (data consumers). Considering the huge amount of data that a single connected car can produce it is not feasible to deploy a centralized Cloud-based solution that stores all the captured data, to make it available to external applications or services. A more sustainable and realistic approach would involve a platform for live-data delivery. In this case, the third party would be subscribed to certain dataflows, and it would decide what to do with the incoming data (process it, store it or discard it).

The vehicle data exploitation platforms would need to include functions for data management, data monetization and cybersecurity as well as providing data access mechanisms based either on human-machine interfaces (HMI) or application programming interfaces (API). 5G, Cloud and Multi-access Edge Computing (MEC) are the key enabling technologies (KETs) for this vehicle data exploitation platform. The following figure shows the conceptual architecture of the platform.

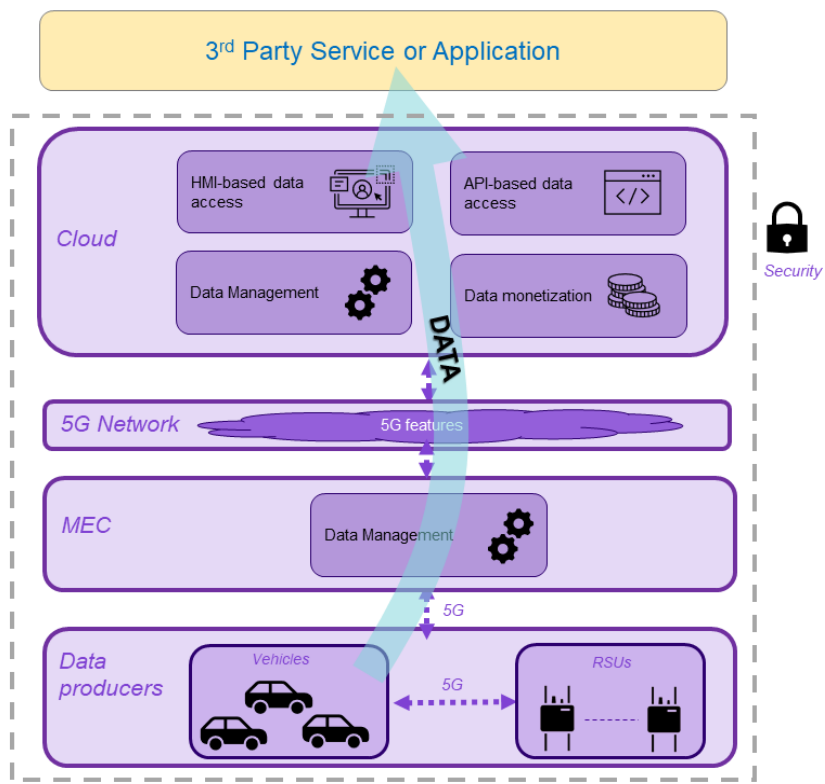

**Figure 1 Conceptual architecture of the vehicle data exploitation platform**

## Vehicular data space

This section describes the vehicular data space that needs to be considered by a vehicle data exploitation solution. The vehicular data space refers to the information about a vehicle and its surrounding space, e.g., road events and environmental conditions. The information can be retrieved directly from a vehicle, from the roadside infrastructure and from other general sources of information, such as a traffic control centre or social media. Data can be static over time if they represent some specific fixed attributes of the vehicle or, conversely, dynamic if they describe some variable characteristics or parameters of vehicles. The categories of vehicular data are summarized in the following table:

**Table 1 – Vehicular data categories**

| Data category | Description |
| --- | --- |
| Vehicle Identification | Data defining the identity of the vehicle. |
| Vehicle Attributes | Data defining the physical characteristics of the vehicle that do not change over time. |
| Vehicle Status | Data related to the current status of the vehicle gathered by sensors installed on-board of the vehicle. |
| Vehicle Position and | Data providing information about the position and the dynamics of |

| Dynamics | the vehicle. |
|---|---|
| Sensors' vehicle external data | Data provided by sensors that are installed on-board of the vehicle and providing data not related to the vehicle itself, e.g. obstacle, road surface condition. |
| Trip information | Data related to the current trip of the vehicle. |
| Driver status and information | Data related to the driver, data can be either dynamic data such as the health status or static data such as driver preferences for some specific service. |
| Roadside infrastructure | Data available from the roadside infrastructure. |
| External information | Information related to cyber-physical human-centric activities (e.g., local events calendar, news, weather forecast, social media) that can be retrieved from sources that are external to the vehicle and to the roadside infrastructure. |

## 5G features and standards

### Software Defined Networking (SDN)

SDN is one of the key technologies for 5G networks, together with Network Function Virtualization (NFV), as they allow network virtualisation, needed to provide the flexibility required for 5G network architecture, in opposition to previous static architectures. The concept of SDN has been firstly introduced in 3GPP TS 23.501, where the overall 5G system architecture is described. SDN centralizes the intelligence of the network through software-based controllers (SDN controllers), which have a global view of the network. The advantages of SDN are:

- Centralised Provisioning managing devices of several vendors using standard protocols
- High scalability and flexibility to increase or decrease network resources
- Definition and distribution of centralized security policies
- Reduced hardware footprint
- QoS improvement

### Network Function Virtualization (NFV)

The concept behind Network Functions Virtualization (NFV) is to create a network architecture that uses technologies widely adopted in IT virtualization, to virtualize all the existing (and new) network node functions.

This architecture is standardized by the Network Functions Virtualization (NFV) Industry Specification Group (ISG) created by the European Telecommunications Standards Institute (ETSI). In detail, NFV 001 defines the main NFV use cases while NFV 002 depicts the general architecture. The main NFV concept and the related terminology has been reported in NFV 003. Other important definitions are described in NVF 004 and NFV-INF 001, that respectively define the Virtualization requirements and the architecture of the NFV Infrastructure (NFVI) which supports deployment and execution of Virtualised Network Functions (VNFs). Further aspects related to the NFV architecture, its interfaces and related requirements are specified in the ETSI NFV-IFA set of standards. Security and trust aspects of NFV are specified in the set of ETSI NFV-SEC standards. Representations for implementation of ETSI NFV-IFA and ETSI NFV-SEC set of standards are specified in the ETSI NFV-SOL standards where protocol and data models have been specified providing possible solutions (i.e., SOL stands for Solutions) for the implementations of REST based APIs for the interfaces at the different reference points.

*Network Slicing*

5G networks need to support a diverse list of verticals such as automotive, healthcare, energy, manufacturing, infotainment, etc., with an exhaustive number of use cases each. These use cases demand very different requirements in terms of network performance that cannot be provided by 4G (and previous) networks. The solution is realized through network slicing. A logical network, named as a 5G network slice, is composed of several elements of the network: access network-radio or wired-, core network, transport, edge network elements, etc. Each of these elements may be used in a dedicated way or be shared with other network slices. At the same time, these resources may belong to different operators or even other network suppliers. Each 5G network slice is isolated from the other network slices, in a way that the use of the other slices will not affect the QoS provided by this slice.

Network slicing has been introduced in 3GPP TS 23.501. The management of network slicing is examined in 3GPP TS 28.530 where concepts of network slicing together with main use cases and requirements are detailed. The provisioning of network slicing for 5G networks and services is instead specified in 3GPP TS 28.531.

*Multi-access Edge Computing (MEC)*

Multi-access Edge Computing (MEC) is a new paradigm that provides computing, storage, and in general networking resources within the edge of the network. While at the beginning the group was focused on cellular networks, today it has broadened its interest for every network type, e.g., Wi-Fi or cabled. MEC servers are deployed on a generic computing hardware, and allow for delay-sensitive, highly scalable, and context-aware applications to be executed in close proximity to the end users. An application can also be a network function to be chained with others to realize a more complex end-to-end service.

MEC standardization activities are held in a dedicated ETSI ISG group. Several standards have been released, the main one is: MEC 001 which provides a glossary of terms relating to the conceptual, architectural, and functional elements within the scope of Multi-access Edge Computing. On the other hand, MEC 002 specifies the requirements for Multi-access Edge Computing with the aim of promoting interoperability and deployments. The framework and reference architecture for MEC are described in MEC 003 where a MEC system that enables MEC applications to run efficiently and seamlessly in a multi-access network is depicted.

Moving from the architecture to the APIs, it is worth mentioning MEC 009, which defines the design principles for RESTful MEC service APIs, provides guidelines and templates for the documentation of these, and defines patterns of how MEC service APIs use RESTful principles. The lifecycle of the applications together with rules and requirements management are highlighted in MEC 010. Other APIs relevant to 5GMETA are presented in MEC 013 that describes the Location APIs, useful to access the UE position as known by the network.

MEC groups are also working on the CCAM field: MEC 022 is a Study on MEC Support for V2X Use Cases, while MEC 030 specifies the API to facilitate V2X interoperability in a multi-vendor, multi-network, and multi-access environment.

*Cybersecurity*

5G networks are subject to many threats which could be exploited by several threat agents such as insiders, nation states, cyber-warriors, hacktivists, corporations, cyber-terrorists, and scripts-kiddies. 3GPP TS 33 501 defines different sets of security measures organised by domains:

- Network access security: the set of security features that enables a UE to authenticate and access services via the network securely, including the 3GPP access and Non-3GPP access, and in particularly, to protect against attacks on the (radio) interfaces.

- Network domain security: the set of security features that enable network nodes to securely exchange signalling data and user plane data;

- User domain security: the set of security features that secure the user access to mobile equipment;

- Application domain security: the set of security features that enable applications in the user domain and in the provider domain to exchange messages securely.

- Service Based Architecture (SBA) domain security: the set of security features that enables network functions of the SBA architecture to securely communicate within the serving network domain and with other network domains. Such features include network function registration, discovery, and authorization security aspects, as well as the protection for the service-based interfaces.

- Visibility and configurability of security: the set of features that enable the user to be informed on whether a security feature is in operation or not.

*Predictive QoS*

To support Predictive QoS, the network produces and sends to the appropriate V2X services/applications in-advance QoS Notifications (IQN) messages that contain all information required to adapt and react to the foreseen changes in terms of QoS. This 5G feature requires substantial efforts from SDOs to reach a common understanding and framework for Predictive QoS. The 5GAA has initiated this during recent years other SDOs like 3GPP and ETSI have developed various standard to support it, like 3GPP TS 23.288 that specifies 5G Architecture enhancements to support network data analytics services, ETSI GS MEC 22 V2.1.1 that identifies key issue/problematic linked to Predictive QoS notifications or ETSI GS MEC 30 V2.1.1 that focuses on how IQN messages shall be generated by the V2X Information Service and then consumed by Vehicular Application. Finally, this 5G enabler is gaining more and more attention by various European Research Projects like for instance 5GCroco which has identified it as one of its main Key Technology Enablers.

*Precise positioning*

In the 3GPP Rel-16 a work item was dedicated to the specifications of the positioning solutions that the NG-RAN shall support. The 3GPP TS 38.305 specifies the positioning functions of the UEs that are exploited for the computation of the position. The architecture of the location service system is firstly introduced in the technical specification devoted to the overall 5G system architecture. The Location Service is not a compulsory feature at the moment. If it is available, it can be exploited from both commercial and regulatory services.

In 3GPP TS 38.305, the UE Positioning architecture is detailed specifying the operations and the information flows between the end user and the NG-RAN. The information flows, together with the signalling protocols to be used, are strictly dependent on the positioning method that is used.

*C-V2X*

C-V2X (Cellular Vehicle to Everything) is a quite new concept that refers to cellular technologies adapted for connected vehicles communication. C –cellular- refers to 4G LTE and 5G NR 3GPP technologies. C-V2X term covers communication between vehicles, and other road users or entities by using the cellular network (so called, Vehicle to Network –V2N- communication) but also direct communication (either Vehicle to Vehicle -V2V-, Vehicle to Infrastructure –V2I- or Vehicle to Pedestrians -V2P).

From a standards point of view, Release 14 introduced the first enhancements for V2X communications to cellular technologies, in this case for LTE and paying special attention on V2V communications, through the LTE network (Uu interface) and with a direct communication through the PC5 interface. The targeted services are safety, traffic efficiency and infotainment. An application layer security is supported.

Release 15 starts dealing with 5G, specifying the support for data connectivity over NR and LTE access technologies (5G phase 1). It improves PC5 communication and includes support for 64-QAM. The enhanced V2X services foreseen are advanced driving, extended sensors, remote driving, and platooning. Release 16 provides full support for LTE and NR communication including the 5G core (5GC), including interworking between both technologies, together with an extensive list of enhancements such as simultaneous LTE PC5 and NR PC5.

*Mapping of 5G standards with introduced features*

In the following table a compendium of the main 5G standards and technical reports is provided for each of the considered features. This compendium does not aim to be an exhaustive list of all standards, but it aims to identify the main ones that are the most relevant for vehicle data exploitation.

**Table 2 – Main 5G standards and technical reports**

| 5G feature | 5G standards and technical reports |
|---|---|
| SDN | 3GPP TS 23.501 [2], ITU-T Y.3300 [3], ONF TR-502 [4], ONF TR-521 [5], ITU-T Y.3351 [6], ITU-T Y.3355 [7]. |
| NFV | NFV-IFA [8], NFV-SEC [9], NFV-SOL [10], NFV-SWA [11], NFV-MAN [12] |
| Network slicing | 3GPP TS 23.501 [2]. 3GPP TS 28.530 [13], 3GPP TS 28.531 [14] |
| MEC | ETSI GS MEC 001 [15], ETSI GS MEC 002 [16], ETSI GS MEC 003 [17], ETSI GS MEC 009 [18], ETSI GS MEC 010 [19], ETSI GS MEC 013 [20], ETSI GS MEC 022 [21], ETSI GS MEC 030 [22] |
| Cybersecurity | 3GPP TS 33 501 [23], 3GPP TS 33.401 [24], 3GPP TS 23.502 [25] |
| Predictive QoS | 3GPP TS 23.288 [26] |
| Precise Positioning | 3GPP TS 38.305 [27], 3GPP TR 38.855 [28] |
| C-V2X | 3GPP TS 23.303 [29], ETSI TR 102 638 [30], 3GPP TS 22.185 [31], 3GPP TS 22.186 [32] |

**Relevance of 5G features for vehicle data exploitation**

As the main dataflow when providing vehicle data to third parties involves data been moved from vehicles to cloud services, the C-V2X sidelink communications do not became essential, being restrained to their utilisation for communications tethering. Furthermore, the low bandwidth provided by the PC5 interface limits its utilization to some data types and sampling rates. However, the C-V2X feature can be considered to cover legacy CCAM services that are typically implemented in the automotive arena.

A platform that provides vehicle data to third parties requires to have a high degree of flexibility to support the reception and the management of data that are consumed by different Cooperative Connected and Automated Mobility (CCAM) applications. It is expected to deliver different dataflows for different services concurrently and in a scalable way. To this extent, the platform needs to be based on the SDN and NFV approaches. These two 5G features enable to dynamically configure the data platform according to the data processing pipelines that have to be deployed to connect the data producers and data-based CCAM applications in a secure and efficient manner. Furthermore, the definition and application of policies to apply at the SDN, when cost/performance conflicts arise, are key to make the platform meet customer expectations and application requirements while satisfying business models.

The SDN and the NFV features are also the main components to enable network slicing. The platform needs to allocate adequate resources for each data pipeline to fulfil heterogeneous SLAs. The network slicing guarantee each CCAM application to achieve the required performance from data pipelines. Moreover, the business model of the platform should take into account the cost, requested by the network, for allocating the resources needed to achieve the performance requested by the platform's users.

Other aspects that the data platform should deal with are the scalability and low latency communications. The MEC is the essential 5G feature to solve these issues. Aggregating data at the edge of the network allows to handle large volume of data and of data sources. The platform can exploit the MEC feature to distribute the load of some of its functionalities from the cloud platform to the edge servers, while keeping centralized at the cloud level the provision of data for the CCAM applications. Low latency communications are of interest for some CCAM applications that need to provide information to the vehicles. These applications can exploit the availability of computing and hosting resources at the edge servers to decrease the latency of the data communications for CCAM application services requiring prompter processing.

Concerning the low latency CCAM services, the Predictive QoS is also arelevant features to be considered. The Predictive QoS can make it possible to meet the business and performance trade-off and to ensure SLAs. As the volume of data gets higher, the concurrency level increases and the available assets get busy or even congested, these technologies can make the difference to maintain a smooth and seamless dataflow. However, the Predictive QoS feature is at an early stage of development, and it is not yet available in commercial 5G networks.

The data platform also must consider cybersecurity issues in order to ensure the highest security of data delivered to the CCAM applications. This becomes more critical when different systems and network domains. Many of these aspects will be also tackled at the application layer when designing the architecture of the platform. In particular, privacy aspects should be well considered by the platform to guarantee the anonymity of data provided to the CCAM services.

Precise positioning feature is interesting for the scope of the data platform. However, the precise position of UEs is still in an early phase of development, and it is not yet mature enough for a practical use. Furthermore, most of UEs can provide GNSS-based position that is sufficient for the currently implemented solutions. Network positioning can also be used as a cybersecurity feature (or to detect fault) because allows the network to understand if the positions sent by a certain UE can be trusted.

A wrap-up of the previous analysis is provided in the following table.

**Table 3 – Relevance of 5G features for vehicle data exploitation**

| 5G feature | Relevance | Justification |
| --- | --- | --- |
| C-V2X | Medium/Low | The direct communication among vehicles provided by the PC5 interface is not in the main target of a Cloud data platform. Nevertheless, the general architecture (with the use of the Uu interface) is relevant. |

| SDN | Medium | Important to fulfil the needed performance but typically under the control of the MNOs with no possibility to be influenced |
|---|---|---|
| NFV | High | One of the main components of the data platform infrastructure. VNF can host the architecture components, manage the employed resources dynamically to respond SLAs and even host services of CCAM applications developed on top of the data platform |
| Network slicing | High | Pivotal to achieve the performances needed for most of the relevant use cases and isolate dataflows of different CCAM applications |
| MEC | High | Very relevant for all the use cases that need scalability and low latency |
| Cybersecurity | High | Pivotal for every possible use case, including privacy mechanisms tailored to produced data types |
| Predictive QoS | Medium/Low | Interesting and highly relevant but currently not enough mature |
| Precise Positioning | Low | Still in an early phase. Not high relevant as vehicles usually include accurate geoposition systems to label produced data. The usage of GNSS-based systems seems yet to be the best solution |

## Conclusions

5G is a key enabling technology of CCAM and includes several features that can contribute to create a scalable, flexible, reliable and secure data platform that delivers vehicle data to third parties in the Cloud. This paper has identified NFV, Network slicing, MEC and cybersecurity as the main 5G features that will enable such a platform.

However, not all the features introduced by the standards will be available at the same time, but the other way around, the 5G network (at least in Europe) is being built step-by-step. This process has started from the adoption of the 5G new radio with the 4G EPC core (i.e., the so-called NSA) and it will evolve the core network towards the full delivery of all the new characteristics (i.e., SA). The current status of 5G deployment in Europe which is based mainly on NSA, allows the creation of data platforms that can be incrementally enhanced with the future 5G releases.

## Acknowledgements

This work is a part of the 5GMETA project. This project has received funding from the European Union's Horizon 2020 research and innovation programme under grant agreement No 957360. Content reflects only the authors' view and European Commission is not responsible for any use that may be made of the information it contains.